**Assessing the Variety of a Concept Space Using an Unbiased Estimate of Rao's Quadratic Index**

Anubhab Majumder[1*], Ujjwal Pal[1], Amaresh Chakrabarti[1]

[1]*Department of Design and Manufacturing, Indian Institute of Science, Bengaluru, India*

**Corresponding author email: anubhabm@iisc.ac.in*

**Abstract.** Past research relates design creativity to 'divergent thinking', i.e., how well the concept space is explored during the early phase of design. Researchers have argued that generating several concepts would increase the chances of producing better design solutions. 'Variety' is one of the parameters by which one can quantify the breadth of a concept space explored by the designers. It is useful to assess variety at the conceptual design stage because, at this stage, designers have the freedom to explore different solution principles so as to satisfy a design problem with substantially novel concepts. This article elaborates on and critically examines the existing variety metrics from the engineering design literature, discussing their limitations. A new distance-based variety metric is proposed, along with a prescriptive framework to support the assessment process. The framework measures the real-valued distance between two design concepts using any chosen representation of their underlying abstraction levels. The proposed framework is implemented in a software tool called 'VariAnT.' Furthermore, the tool's application is demonstrated through an illustrative example.

## 1 Introduction

Designing is regarded as a means of changing existing situations into preferred ones. The engineering design process has been broadly classified into four stages: task clarification, conceptual design, embodiment design and detail design (Pahl & Beitz, 1996). Conceptual design is an early stage in the design process, which involves generating solution concepts to satisfy the functional requirements of a design problem (Chakrabarti & Bligh, 1994). The conceptual design stage is crucial because a high percentage of the product cost is committed at this stage (Saravi et al., 2008). Decisions made during this stage will strongly affect all the subsequent stages of the design process. Past research (Baer, 2014) relates creativity to 'divergent thinking', i.e., how well the concept space is explored during the early phase of design. The ability to explore the breadth of the concept space is directly related to the ability to restructure problems and is, therefore, an important measure of creativity in design (Shah et al., 2003). Concept space exploration depends on the capacity to produce a wider 'variety' of ideas with higher 'fluency' (Shah et al., 2003). Researchers (Chakrabarti & Bligh, 1994; Shah et al., 2003) have argued that generating several concepts would increase the chances of producing better design solutions, a view that aligns with Osborn's (2012) foundational brainstorming principle that "quantity breeds quality," whereby generating many ideas enhances the chances of arriving at more effective and innovative solutions. Past empirical studies also showed that the variety of the concept space is positively correlated with the novelty of the generated set of ideas (Jagtap et al., 2015; Srinivasan & Chakrabarti, 2010a; Kurtoglu, 2009). Generating many concepts, resulting in higher 'fluency' (i.e., the total count of the number of concepts generated), that differ from one another only in minor or superficial ways does not prove effective in concept generation (Shah et al., 2003). Here, minor or superficial ways refers to differences that do not meaningfully alter the functional, behavioural, or structural aspects of a design concept, for example, changing the colour, naming, or trivial features while maintaining essentially the same underlying principles. Thus, it is imperative to generate and explore a diverse set of alternative solution concepts during the early stages of the design process. This approach ensures that designers have a multitude of options to consider which increases the likelihood that designers can identify and pursue a promising solution.

Variety indicates how well one has explored the concept space of a design problem. Variety metrics, often referred to as 'diversity' metrics, are also popular in other domains such as economics and ecology. Mathematically, a variety or diversity measure should tell us the probability that two objects (or species - in the case of ecology) selected at random (without replacement) from a sample will belong to different groups (Hurlbert 1971; Ahmed et al. 2021).

Other than evaluating concept generation, assessing variety also offers value in design-by-analogy, where designers are presented with external stimuli to inspire new ideas. Prior research has shown that the diversity of stimuli - often captured as the conceptual "distance" between the target design problem and the analogical sources - substantially influences the ideation process, affecting the originality or novelty of resulting concepts (Goncalves et al., 2012; Lu et al., 2023). Therefore, having a computational method to quantify variety becomes useful not only for evaluating concept sets but also for systematically selecting and presenting diverse stimuli to designers, ensuring exposure to analogies that span a broad conceptual range.

This article critically examines the existing variety metrics from the engineering design literature and highlights the limitations of the existing metrics through test cases. Specifically, existing metrics rely on a genealogy tree-based representation of concept space, which assumes that concepts can always be classified into discrete nodes at predefined abstraction levels (physical principle, working principle, embodiment, and detail). This binary, node-based approach limits flexibility, especially when real-valued distances – such as semantic similarity scores – are more appropriate for capturing the relationship between ideas generated at a particular abstraction level, making current metrics unsuitable for such scenarios. Furthermore, some existing metrics demonstrate a lack of accuracy and sensitivity in special test cases, ultimately leading to misleading variety scores that do not adhere to the mathematical definition of variety. Additionally, so far, no software tools are reported in the literature that can automate the variety assessment process directly from textual data of a design concept space. Finally, researchers often report difficulties in clearly defining physical and working principles during genealogy tree creation, indicating a need for a more flexible and robust framework that can accommodate customisable abstraction levels tailored to different design problems. To address these limitations, a new variety metric is proposed based on Rao's quadratic diversity index (Rao 1982), along with a prescriptive framework – embodied into a software tool – to support the assessment process. The proposed metric offers advantages over the existing variety metrics as it eliminates the need for a tree-like representation of a design concept space. This framework uses automated text to vector embedding to measure the real-valued distance between two design concepts across the distinct levels of abstraction. Although the example implementation of the proposed framework uses a particular abstraction scheme, the framework itself is generalisable and can accommodate any representation of abstraction and any distance-assessment method, allowing designers the flexibility to define abstraction levels that best suit their specific design context. The framework is implemented in a software tool called 'VariAnT.' The general applicability of the proposed framework is demonstrated through multiple examples.

## 2 Background and Related Work

This section reviews existing metrics for measuring the variety of a concept space proposed in the engineering design domain. To elaborate on these metrics, Section 2.1 first introduces an example concept space, followed by the construction of its genealogy tree. Then, in Section 2.2, we use this genealogy tree to explain the existing variety metrics in detail.

### 2.1 An example concept space and its genealogy tree

At the conceptual design stage, a designer moves freely between different levels of abstraction and generates ideas corresponding to a particular abstraction level (Srinivasan & Chakrabarti, 2010b). Most of the existing literature on variety considers four abstraction levels: physical principle, working principle, embodiment and detail (Shah et al. 2003, Nelson et al. 2009, Verhaegen et al. 2013, Ahmed et al. 2021, Henderson et al. 2017). Let us consider an example case illustrated in **Figure 1**, where a designer explores ideas with an AI chatbot to fulfil the function of 'pumping water.' The chatbot provided two ideas at the physical principle level: 'centrifugal force' and 'positive displacement.' Now, based on these two physical principles, five concepts are generated, each representing a distinct idea at the working principle level. The exploration can go further at the embodiment and detail level, which may lead to a concept space $\mathrm{C} = \{\mathrm{C}_1, \mathrm{C}_2, \ldots, \mathrm{C}_\mathrm{N}\}$ with $N$ number of concepts where each can be represented in $\alpha$ levels of abstraction. For convenience, we have only represented the concepts at two levels of abstraction: physical principle ($\alpha = 1$) and working principle ($\alpha = 2$). At each $\alpha$, there exists a different set of ideas, $I^\alpha = \{I_i^\alpha \mid i \in 1,2, \ldots, \beta_\alpha\}$, where $\beta_\alpha$ denotes a total number of ideas at $\alpha$. Therefore, a concept $C_i$ can be generated by combining different ideas taken from the idea space $I^\alpha$ as shown in **Figure 1**. We refer to the concept space shown in **Figure 1** as $C^A$ where $N = 5$.

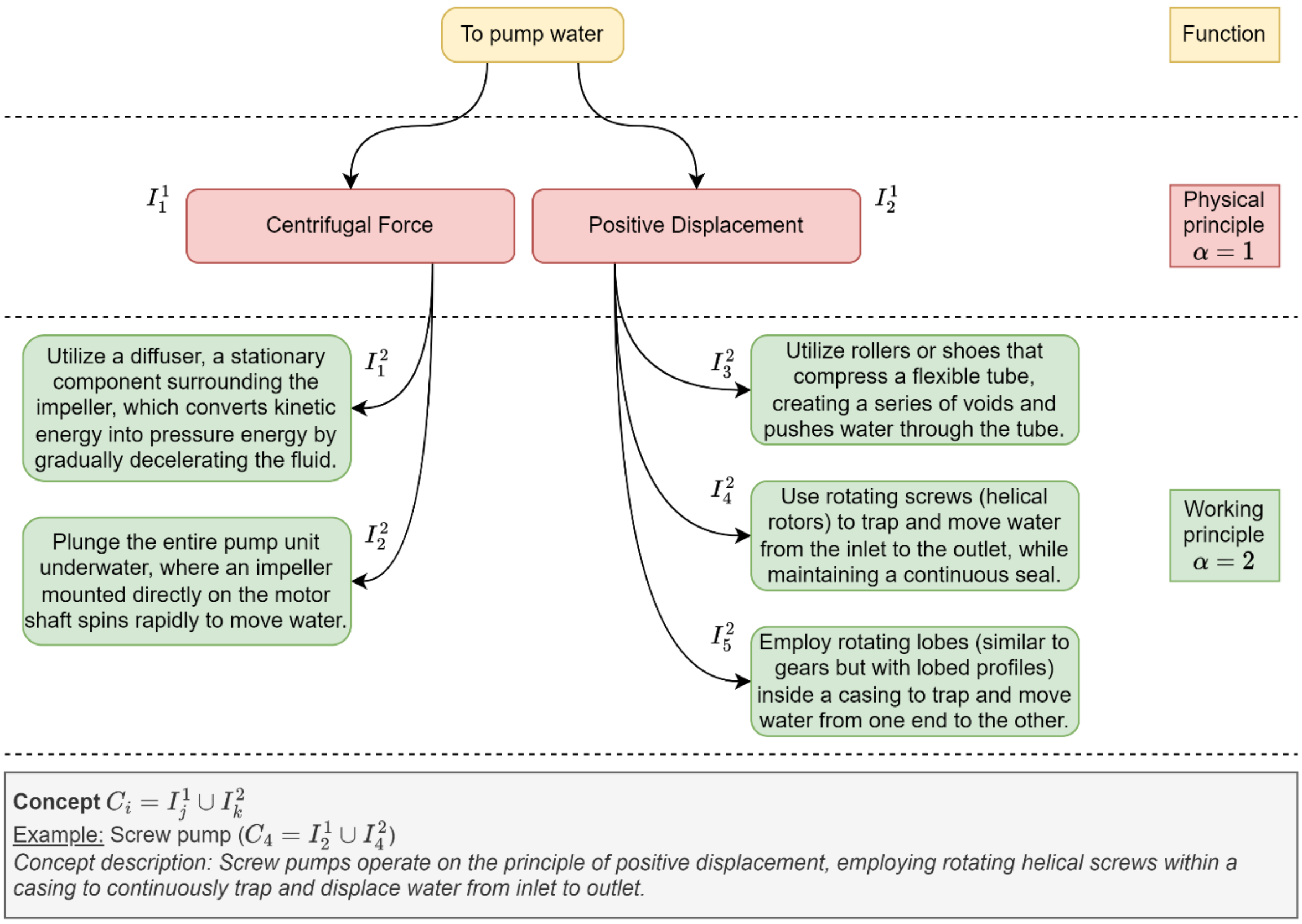

**Figure 1.** Concept space ($C^A$) generated by combining different ideas from different levels of abstraction.

In most of the literature on variety assessment, researchers have used a tree-like structure, called a genealogy tree, to represent a concept space. For example, **Figure 2** illustrates the genealogy tree of concept space, $C^A$. Here, $n_i^\alpha$ denotes the total number of concepts that use the $i^{th}$ idea from the idea space $I^\alpha$. If $\beta_\alpha$ is the maximum number of ideas in $I^\alpha$, then we can write, $\sum_{i=1}^{\beta_\alpha} n_i^\alpha = N$. It is important to note that even if we describe the concept space $C^A$ at the embodiment and detail abstraction level, the genealogy tree will still have 5 nodes at those lower hierarchical levels, with each node containing one idea, and hence the variety score at the embodiment and detail level will be the same as the working principle level, irrespective of the existing variety metrics (elaborated in Section 2.2) that we use. Hence, Using the concept space $C^A$ and its corresponding tree as a common ground; we have elaborated on the existing variety assessment techniques in the following part of this section.

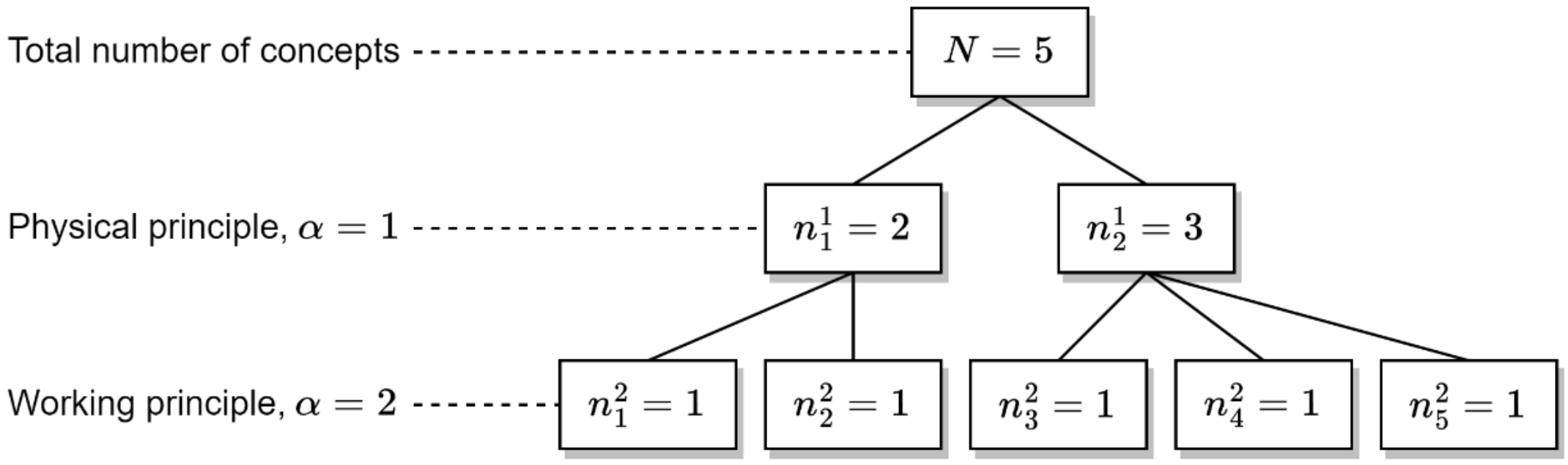

**Figure 2.** Tree constructed from concept space $C^A$.

## 2.2 Existing Variety Metrics in Engineering Design

In engineering design, there are two commonly used approaches to assess design variety: subjective ratings and objective ratings. As an example of subjective evaluation, Linsey et al. (2011) proposed a method where a coder intuitively categorises design concepts based on their overall differences. Concepts with similar features are sorted into different bins or pools. At the end of the sorting process, an individual's variety score is calculated as the ratio of the number of bins the concepts were sorted into to the total number of bins. This subjective approach relies on the coder's mental model rather than a numerical procedure (Ahmed et al., 2021). In contrast, the objective approaches replace subjective human raters with a deterministic formula that relies on a few measured attributes of a set of designs. Our work particularly focuses on objective approaches. Here, we have reviewed four existing objective variety metrics related to the engineering design domain: the variety metrics proposed by Shah et al. (2003), Nelson et al. (2009), Verhaegen et al. (2013), and Ahmed et al. (2021). Note that other objective variety metrics also do exist in engineering design literature, but we kept those out of our discussion for the following reasons: (a) for example, the metric proposed by Henderson et al. (2017) uses other design assessment metrics such as quality, effectiveness, novelty, applicability, etc. to calculate the variety of the solution concepts and thus, the variety metric cannot be considered as an independent measure; (b) in case of the metric proposed by Srinivasan and Chakrabarti (2009), the measure is dependent on the order in which the ideas are generated, even though this should not have any impact of the overall variety of a concept space as long as it contains the same set of concepts, and hence the metric is not applicable for the concept space $C^A$.

### 2.2.1 Metric proposed by Shah

Shah et al. (2003) proposed a metric for measuring the variety of a concept space as follows. The design problem is first decomposed into its essential functions or characteristics. The conceptual origins (i.e., physical principles, working principles, embodiment, and details) of the concepts are analysed through hierarchical or abstraction levels based on how the concepts fulfil each design function. At the highest level of abstraction ($\alpha = 1$), concepts are differentiated by the physical principles used by each to satisfy the same function; this is the most significant extent of finding differences between concepts. At the second level ($\alpha = 2$), concepts are differentiated based on working principles, even though they share the same physical principle. At the third ($\alpha = 3$) and fourth ($\alpha = 4$) levels, concepts have different embodiment and detail, respectively. The number of branches in the genealogy tree indicates the variety of concepts. If greater variety is to be valued, branches at upper levels should get a higher rating than the number of branches at lower levels. Shah et al. (2003) have assigned values of 10, 6, 3, and 1 to physical principle, working principle, embodiment, and detail levels, respectively. If there is only one branch at a given level (i.e., $\beta_\alpha = 1$), it shows no variety, and the score assigned is 0; otherwise, the score is the number of branches times the weight assigned to that level. A genealogy tree needs to be constructed for each function of a concept. Not all functions are equally important, so a weight $f_j$ is assigned to account for the importance of each. Then, the overall variety measure $V$ takes the following form:

$$V = 10 \times \sum_{j=1}^{m} f_j \sum_{\alpha=1}^{4} \frac{w_\alpha \beta_\alpha}{V_{max}} \quad (1)$$

In Equation 1, $\beta_\alpha$ is the total no. of branches at level $\alpha$; $w_\alpha$ is the weight for level $\alpha$ (suggested weights are: $w_1 = 10, w_2 = 6, w_3 = 3$, and $w_4 = 1$); $m$ is the total no. of functions; and $V_{max}$ is the maximum possible variety score. $V_{max}$ would be obtained if all concepts used different physical principles ($\alpha = 1$). Thus, $V_{max} = w_1 \times N$ where $N$ is the total number of concepts. Therefore, Equation 1 reduces to

$$V = \sum_{j=1}^{m} f_j \sum_{\alpha=1}^{4} \frac{w_\alpha \beta_\alpha}{N} \quad (2)$$

Now, applying Equation 2 to the concept space $C^A$ results in variety score $V(C^A) = \frac{(10\times2)+(6\times5)}{5} = 10$.

### 2.2.2 Refinements proposed by Nelson

Nelson et al. (2009) refined the above metric of Shah et al. to resolve the following two limitations: (a) it produces a lower variety score for greater variety by double counting the ideas at each level in the tree, and (b) the variety score is normalised by the number of concepts, which may not indicate the actual design space exploration. Instead of using the number of branches at each level, Nelson et al. use the number of differentiations of branches at a particular level. For example, two branches at the physical principle level correspond to only a single differentiation between physical principles (i.e. there is only one difference), and three branches at the physical principle level correspond to two differentiations, and so on. Thus, the number of differentiations is always one less than the number of branches at a given hierarchical level. No differentiations occur when a single branch emanates from a node. So, Equation 3 is modified as follows:

$$V = \sum_{j=1}^{m} f_j \left( \frac{w_1(\beta_1 - 1) + \sum_{\alpha=2}^{4} w_\alpha \sum_{l=1}^{\beta_{\alpha-1}} d_l}{N-1} \right) \quad (3)$$

Where $w_1(\beta_1 - 1)$ is the score for differentiation at the physical principle level ($\alpha = 1$), $d_l$ is the number of differentiations at node $l$ (one less than the number of branches emanating from node $l$). One is subtracted from $N$ to preserve the normalisation from 0 to 10 since the maximum number of differentiations is one less than the number of concepts. Nelson et al. have also shown that, even when the number of concepts increases, the variety may decrease as it is normalised by the number of concepts, whereas a non-normalised variety score measures the actual design concept space exploration. So, Equation 3 reduces to:

$$V = \sum_{j=1}^{m} f_j \left( w_1(\beta_1 - 1) + \sum_{\alpha=2}^{4} w_\alpha \sum_{l=1}^{\beta_{\alpha-1}} d_l \right) \quad (4)$$

Nelson et al. have suggested the values of $w_\alpha$ to 10, 5, 2, and 1 to assure that at least two concepts at a lower hierarchical level must be added to equal the variety gain by adding a single concept at the next higher hierarchical level. Applying Equation 3 to the concept space $C^A$ results in variety score, $V(C^A) = \frac{10\times(2-1)+5\times(1+2)}{5-1} = 6.25$.

### 2.2.3 Refinements proposed by Verhaegen

It can be observed from Equation 2 and Equation 3 that, in both cases, the variety measure does not consider the distribution of $N$ concepts over $\beta_\alpha$ nodes, i.e., $n_i^\alpha$. Thus, using both metrics results in a similar variety score for the cases where $\beta_\alpha$ remains constant, and the distribution of $N$ concepts in the idea space (i.e., $n_i^\alpha$) differs across $\beta_\alpha$ nodes. Verhaegen et al. (2013) addressed the same issue and termed this as "accounting for the degree of uniformness of distribution". Following the refinement proposed by Verhaegen et al., Equation 2 can be modified into Equation 5 by replacing $\beta_\alpha$ in Equation 2 with the inverse of the Herfindahl index (Herfindahl 1997).

$$V^\alpha = w_\alpha \sum_{j=1}^{m} f_j \left( \frac{1}{N \times H_\alpha} \right), H_\alpha = \sum_{i=1}^{\beta_\alpha} p_i^2 \quad (5)$$

In Equation 5, $V^\alpha$ denotes the variety score at abstraction level $\alpha$, $H_\alpha$ is the Herfindahl index at level $\alpha$; $p_i$ is the proportion of ideas of variable $i$, i.e., for the $i^{th}$ idea in the idea space $I^\alpha$, $p_i = \frac{n_i^\alpha}{N}$.

Using Equation 5, the variety score at $\alpha = 1$ for the concept space $C^A$ can be calculated as $V^1(C^A) = 10 \times (\frac{5}{2^2+3^2}) = 3.85$. Similarly, for $\alpha = 2$, the score $V^2(C^A) = 6 \times (\frac{5}{1^2+1^2+1^2+1^2+1^2}) = 6$. Unlike Shah et al. and Nelson et al., Verhaegen et al. did not provide an aggregated variety score formula considering all abstraction levels. Nonetheless, one approach to derive an overall variety score $V$ is through a weighted average, defined as:

$$V = \frac{\sum V^\alpha}{\sum w_\alpha} \quad (6)$$

Equation 6 yields an overall variety score for concept space $C^A$ in a scale of 0 to 1, $V(C^A) = \frac{3.85+6}{10+6} = 0.615$.

### 2.2.4 Metric proposed by Ahmed

Ahmed et al. (2021) described the Sharma–Mittal entropy (SME) as a generalised class of methods for measuring diversity (or variety) in other 'non-engineering' domains. They have shown that the Herfindahl index (also known as Herfindahl-Hirschman index or HHI), adopted by Verhaegen et al. (2013), can also be derived from SME. However, instead of using the inverse of the Herfindahl index, they have proposed a variant of the HHI named HHID (i.e., Herfindahl-Hirschman Index for Design) as a new metric for measuring variety. According to the proposed method, the variety measure of a concept space at an abstraction level $\alpha$ can be written as follows:

$$V^\alpha = 1 - \frac{\sum_{i=1}^{\beta_\alpha} (n_i^\alpha)^2}{N^2}, \ for\ N, \beta_\alpha \geq 1 \quad (7)$$

Ahmed et al. (2021) have compared the above-proposed metric with other existing variety metrics by Shah et al. (2003) and Nelson et al. (2009). They have shown that the HHID has advantages over the existing metrics regarding accuracy, sensitivity, optimisability and generalisability. Accuracy was measured with respect to ground truth data sets constructed by experts, and it was found that the HHID-based scores align better with human ratings compared to the other two existing measures. Ahmed et al. (2021) proposed an empirical method of evaluating metric sensitivity by randomly selecting sets of different concepts and comparing the scores obtained by different metrics. The results showed that Shah's and Nelson's metrics gave the same scores to a large percentage of sets and thus proved less sensitive compared to HHID. They have also stated that the HHID closely follows the Gini-Simpson Index (GSI), commonly used as a measure of diversity in ecology, given as follows:

$$GSI = 1 - \lambda, \quad \lambda = \sum_{i=1}^{Z} \pi_i^2 \qquad (8)$$

In Equation 8, $\lambda$ is known as the Simpson Index (Simpson, 1949). $\pi_i (i = 1 \dots Z)$ are the proportion of individuals in the various groups in an infinite population where each individual belongs to one of $Z$ groupings. $\lambda$ can be interpreted as the probability that two individuals that are randomly and independently picked from the population belong to the same group. In contrast, the complement of $\lambda$ in Equation 8 equals the probability that the two individuals belong to different groups. This is also known as the probability of inter-species encounter (Hurlbert, 1971).

It is important to note that the HHID given in Equation 7 is defined for a sample concept space with $N$ concepts, where the estimate of $\lambda$, i.e., $\hat{\lambda}$, is considered as $\sum_{i=1}^{\beta_\alpha} \left(\frac{n_i^\alpha}{N}\right)^2$. However, using the unbiased estimate of $\lambda$, given by Simpson (1949), Equation 7 can be modified as follows:

$$V^\alpha = 1 - \frac{\sum_{i=1}^{\beta_\alpha} n_i^\alpha (n_i^\alpha - 1)}{N(N-1)}, \; for \;\; N \geq 2, \beta_\alpha \geq 1 \qquad (9)$$

We may refer to Equation 8 as the Gini-Simpson Index for Design or GSID. Equation 8 is also known as the Gini-Simpson Diversity Index and is widely used by ecologists as a well-known conventional index for measuring diversity in an ecosystem (Chen et al., 2018; Augousti et al., 2021). The unbiased estimate of $\lambda$ used in Equation 8 is also equivalent to the normalised Herfindahl-Hirschman Index (Cracau & Lima, 2016). Now, applying Equation 7 to the concept space $C^A$ results in variety scores $V^1(C^A) = 0.48$ and $V^2(C^A) = 0.8$ at $\alpha = 1$ and $\alpha = 2$, respectively. Whereas Equation 9 yields variety scores $V^1(C^A) = 0.6$ and $V^2(C^A) = 1$ at $\alpha = 1$ and $\alpha = 2$, respectively. For large sample sizes ($N \to \infty$), GSID asymptotically follows the HHID. The advantage of using GSID instead of HHID as a measure for variety in engineering design is further discussed in Section 3.

## 3 Research Gap: Issues with Existing Variety Metrics

To investigate the issues with the aforementioned variety metrics, we consider two essential properties, 'accuracy' and 'sensitivity', that make a variety metric reliable. The properties are discussed as follows. Ahmed et al. (2021) proposed a procedure for empirically estimating the accuracy of a variety metric by comparing its alignment with a ground truth data set. The ground truth data sets were prepared considering expert feedback, domain knowledge, or consensus from many individuals. Here, the property of 'accuracy' implies the validity of a metric. A measure can only be validated against an external frame of reference or a universally accepted standard. Using experts' ratings to ensure the validity of a metric is a common practice in creativity research (Hennessey et al. 1999). HHID, as a metric for measuring design variety (Equation 7), was validated by Ahmed et al. (2021) against two ground truth data sets; one was established by using pairwise comparisons between sets of polygons, and the other was constructed using milk frother design sketches. Experiments were conducted to benchmark the HHID with the commonly used variety metrics given by Shah et al. (2003) and Nelson et al. (2009). Results from the experiments showed that HHID outperforms the other two metrics in terms of 'accuracy.'

However, in this research, we adopted a theoretical approach to evaluate existing metrics in terms of 'accuracy.' We use the mathematical definition of variety employed by Ahmed et al. (2021) to formulate the property of 'accuracy,' as given below.

**Property 1 (Accuracy).** *Considering a particular abstraction level $\alpha$, the variety score provided by a metric should tell us the probability that two concepts selected at random (without replacement) from a sample concept space will belong to different idea groups (i.e., different nodes of a genealogy tree) at abstraction level $\alpha$.*

Apart from 'accuracy,' a metric can also be evaluated in terms of 'sensitivity.' The property of 'sensitivity' is defined as follows.

**Property 2 (Sensitivity).** *In an ideal case, a variety metric should be able to reflect a change in measurement with varying numbers of concepts in a concept space, i.e., $N$, as well as the distribution of concepts over the $\beta_\alpha$ nodes in the idea space $I^\alpha$, denoted by $n_i^\alpha$.*

For example, let us consider another example of concept space, $C^B$, illustrated in **Figure 3**, where a designer explores ideas to fulfil the function of 'pumping water.' The ideas at the physical principle level are identical to $C^A$ (see **Figure 1**): 'centrifugal force' and 'positive displacement.' Five concepts are generated based on these two physical principles, each representing a distinct idea at the working principle level. The genealogy tree of the concept space, $C^B$, is shown in **Figure 4**. Unlike $C^A$, in the case of $C^B$, 4 out of 5 concepts share identical ideas ('positive displacement') at the physical principle level (as shown in **Figure 3**). This results in different $n_i^\alpha$ values for $C^A$ and $C^B$ at the physical principle level, and a 'good' variety metric should reflect this difference while accurately providing the variety scores for these two concept spaces. In ecology, this property implies the 'evenness sensitivity' of a diversity (or variety) measure (Crupi, 2019).

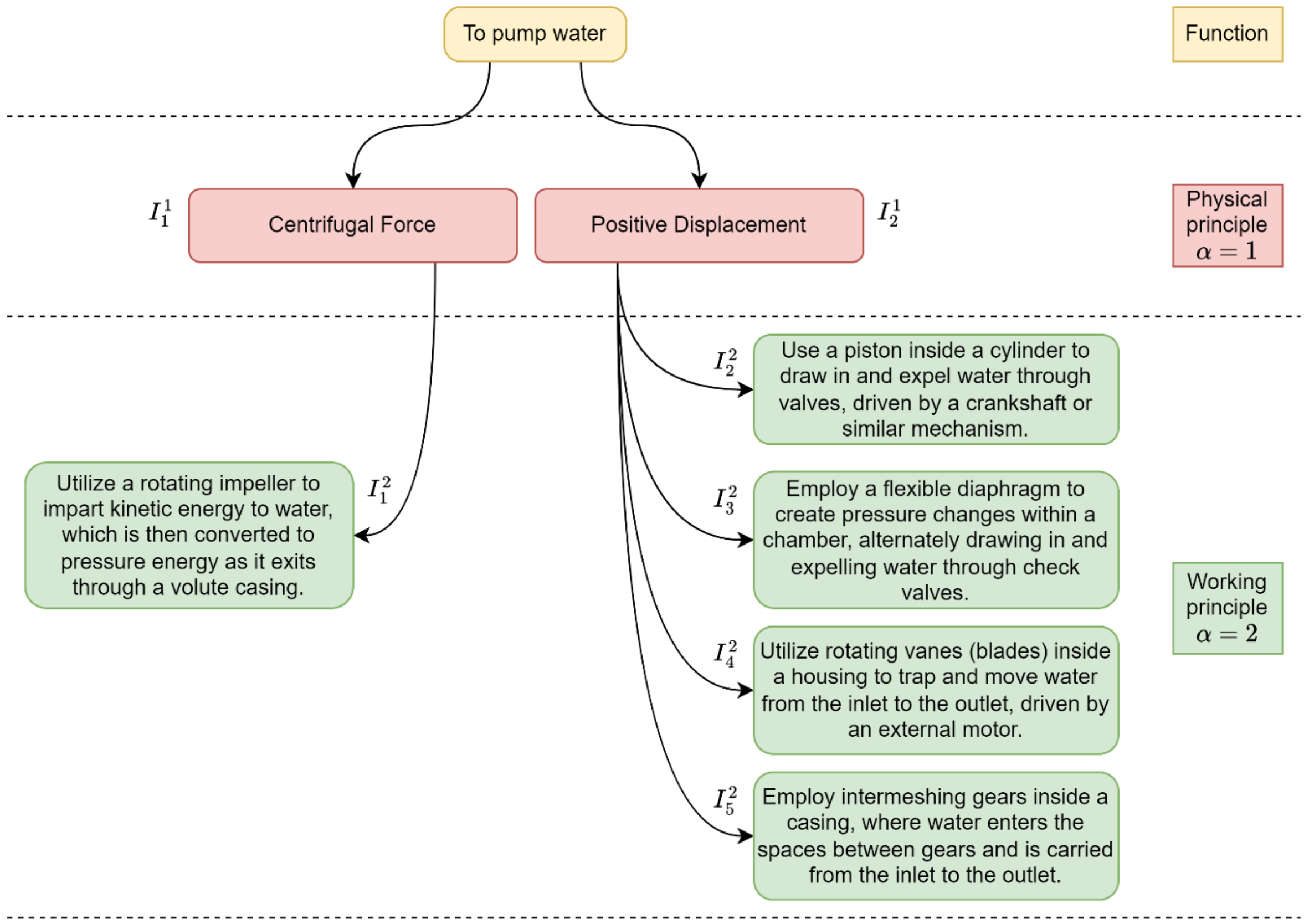

**Figure 3.** Concept space ($C^B$) generated by combining different ideas from different levels of abstraction.

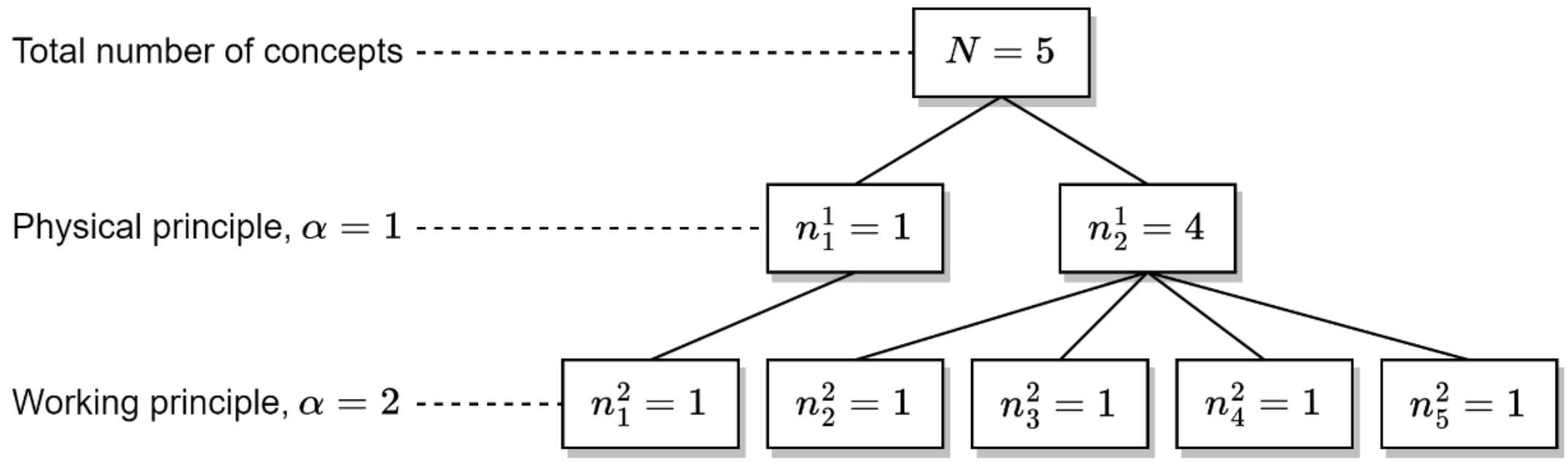

**Figure 4.** Tree constructed from concept space $C^B$.

In the following sections, we follow a theoretical approach to evaluate the accuracy and sensitivity of the existing metrics. We argue that GSID (Equation 9) can be used as a refined version of HHID (Equation 7) to provide better accuracy and sensitivity for genealogy tree-based variety assessment compared to other existing metrics. The argument is supported by at least two test cases where the existing metrics are not found to be accurate or sensitive. The variety metrics proposed by Shah, Nelson, Verhaegen and Ahmed are hereafter referred to as 'SVS', 'NM', 'IHI', and 'HHID', respectively.

### 3.1 Test Case I

Consider a concept space with $N$ number of concepts. At an abstraction level $\alpha$ (e.g., 'physical principle' in **Figure 1**), the total number of ideas or nodes is two, i.e., $\beta_\alpha = 2$ (e.g., 'centrifugal force' and 'positive displacement'). Now, there could exist different concept spaces with an identical number of total concepts, say $N = 20$, but with different distributions of concepts over the two nodes, as shown in **Figure 5**. In this case, adhering to the definition of accuracy and sensitivity, a compelling variety metric should satisfy the following three conditions: (a) It should give the maximum score to a completely even distribution of concepts over the two nodes in the idea space (i.e., $n_1^\alpha = n_2^\alpha = \frac{N}{2}$) as the probability that two concepts that are randomly and independently selected from the concept space share different ideas from the idea space $I^\alpha$ is maximum in this case; (b) A lower variety score should be given for a skewed distribution of concepts over the two nodes in the idea space; and (c) The score at $\alpha$ should be strictly 0 when all the concepts share identical ideas (e.g., all the concepts share a similar physical principle – 'centrifugal force').

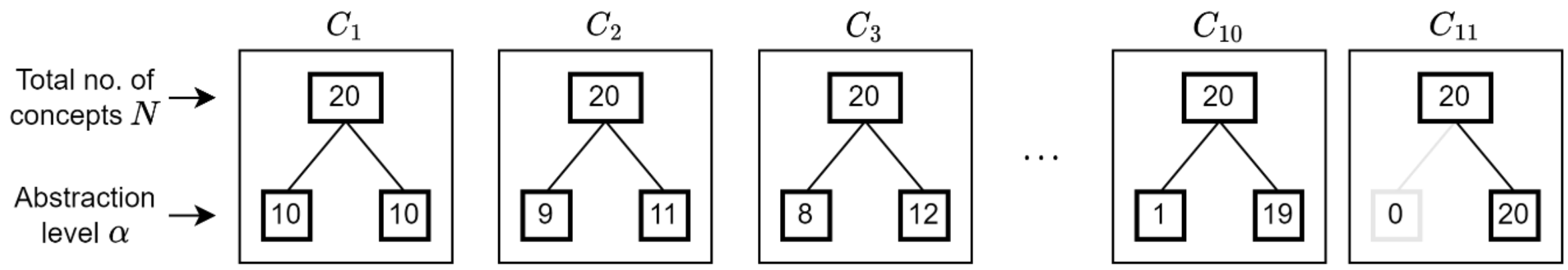

**Figure 5.** Example concept spaces for Test Case I with different distributions of concepts over the two nodes at an abstraction level α.

**Figure 6** presents the scores, scaled to a range of 0 to 1 (with 1 indicating the maximum variety), provided by different variety metrics for different distributions of 20 concepts over two nodes at α. For example, '$5/15$' on the x-axis denotes the distribution of 20 pump concepts at the physical principle level, where five concepts work based on 'centrifugal force', and the remaining 15 concepts work based on 'positive displacement.' As shown in **Figure 6**, the SVS and NM scores remain constant for both even and skewed distributions except for the '$0/20$' distribution, where the score changes from a constant value to 0, and this is due to the change in $\beta_\alpha$ value from 2 to 1. From this observation, it can be concluded that both SVS and NM lack 'sensitivity,' leading to inaccurate variety scores for this test case. In the case of IHI, HHID and GSID, the scores gradually decrease when the distribution of concepts becomes more skewed and thus, these three metrics satisfy the property of 'sensitivity.' However, for the '$0/20$' distribution, unlike the other metrics, IHI gives a score of 0.05 instead of 0. Hence, IHI provides an inaccurate score (violating Property 1) for any finite value of $N$ when $\beta_\alpha = 1$. Overall, for Test Case I, both HHID and GSID are found to be more compelling compared to the others in terms of both 'accuracy' and 'sensitivity.'

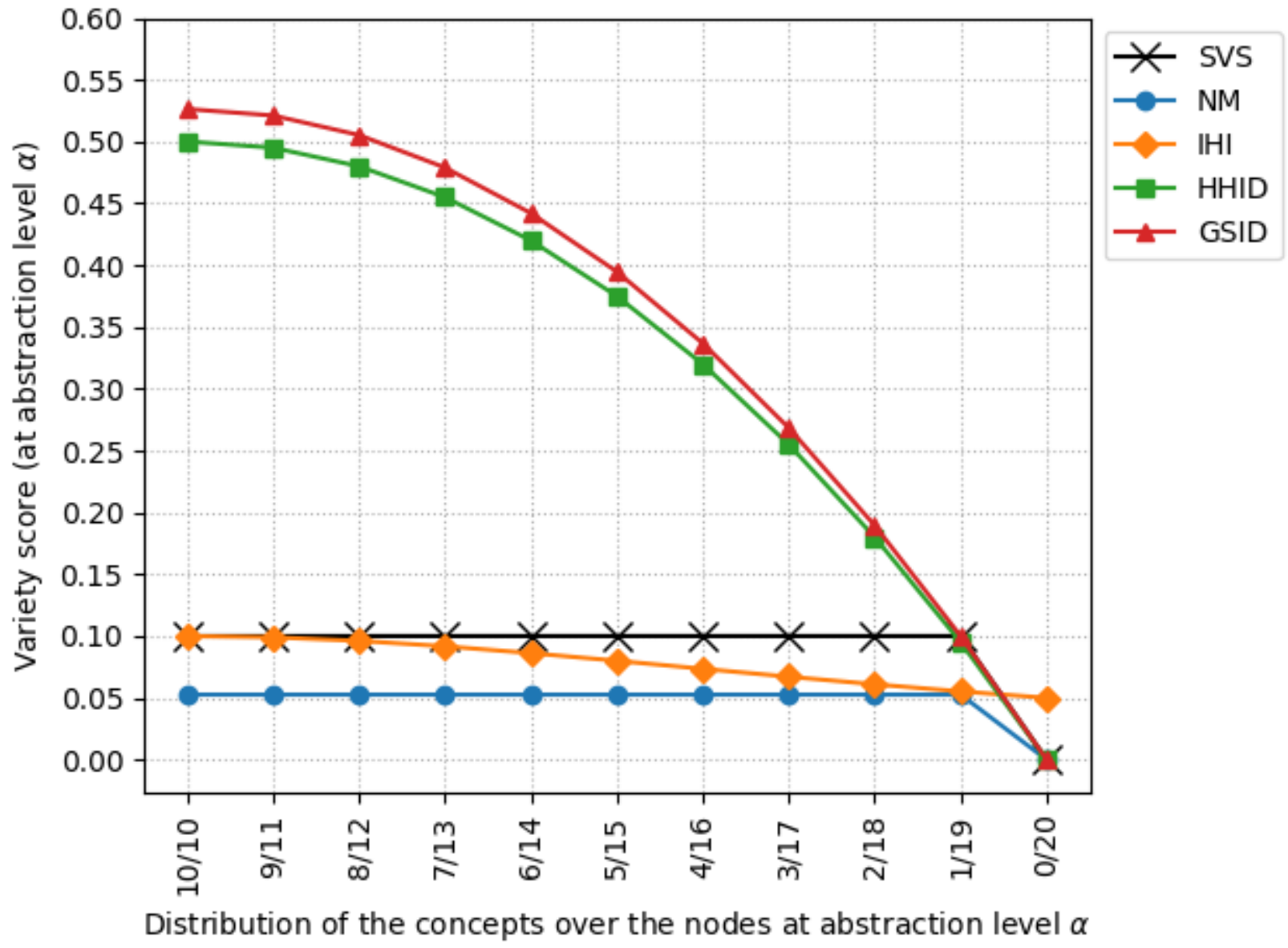

**Figure 6.** Illustration of the Test Case I.

### 3.2 Test Case II

In this test case, we consider different concept spaces, as shown in **Figure 7**, where for each concept space, the concepts are evenly distributed over two nodes in an idea space $I^\alpha$. For example, if there are $N$ pump concepts, $N/2$ of them work based on 'centrifugal force', and the rest $N/2$ work based on 'positive displacement', i.e., $n_1^\alpha = n_2^\alpha = N/2$. Now, consider there are only two concpets in the concept space, i.e., $N = 2$ and both of them have distinct physical principles. In this case, for the abstraction level 'physical principle', adhering to the definition of accuracy (Property 1), the variety score is expected to be 1 (considering the variety metric provides a score in the range 0 to 1, and 1 denoting the maximum variety). If $N$ is large, considering the definition of accuracy (Property 1), the variety score should asymptotically follow the value 0.5, which is the probability of two concepts randomly and independently selected from the concept space (with $N \rightarrow \infty$) are different at the physical principle level.

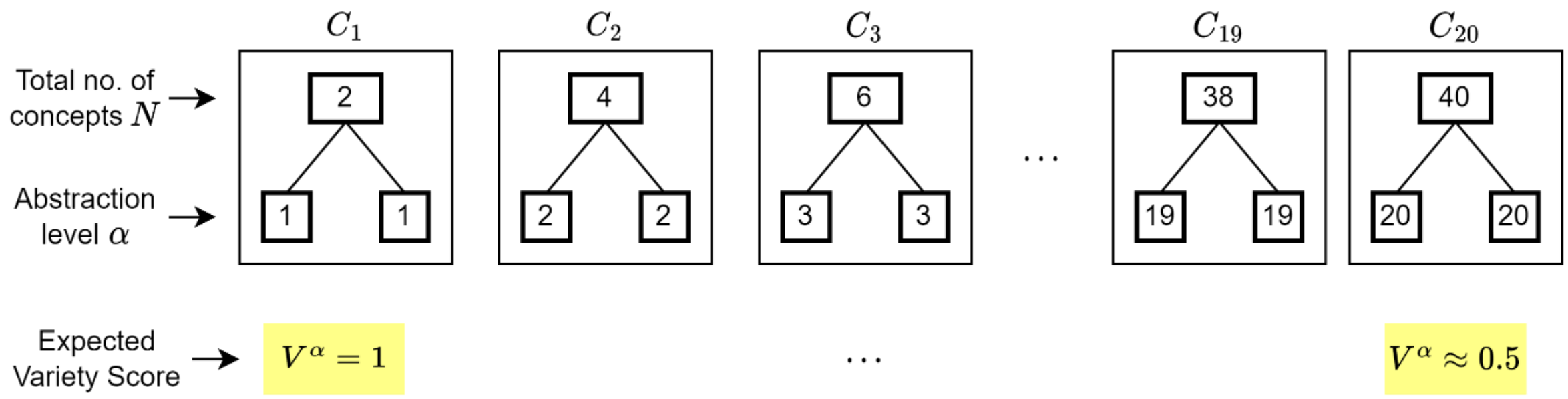

**Figure 7.** Example concept spaces for Test Case II with similar distributions of concepts over the two nodes at an abstraction level α.

Considering the case described above, the scores given by different variety metrics, scaled to a range of 0 to 1, are plotted in **Figure 8**, where $N$ ranges from 2 to 40 (note that the range is chosen arbitrarily). It can be observed that, unlike other metrics, the scores calculated using HHID remain unchanged for varying $N$. HHID also provides an 'inaccurate' score for $N = 2$, i.e., 0.5 instead of 1. It can be noticed that the score, that HHID gives for any finite value of $N$, is nothing but the score obtained from GSID for $N \rightarrow \infty$. This has occurred because of using a 'biased' estimator of the Simpson Index ($\lambda$, see Equation 8) while defining HHID (see Equation 7). The bias is significant when $N$ is small and thus leads to an 'inaccurate' measure of variety, violating Property 1. The other three metrics, i.e., SVS, NM, and IHI, are found to be 'accurate' for $N = 2$. However, when $N$ is large, these three metrics provide variety scores, which asymptotically diminish to the value 0 instead of following the value 0.5 and hence are found to be 'inaccurate,' or violating Property 1. Theoretically, one can also observe from Equation 2 and Equation 3 that the cases when $\beta_\alpha$ is much smaller compared to $N$, both SVS and NM will give a score close to 0 irrespective of the distribution of $N$ concepts over the $\beta_\alpha$ nodes in the idea space. The same is true for IHI as well. For a smaller $\beta_\alpha$ and larger $N$, in Equation 5, the denominator value becomes much greater than the numerator because of the large square terms in the denominator, which ultimately reduces the variety score close to 0.

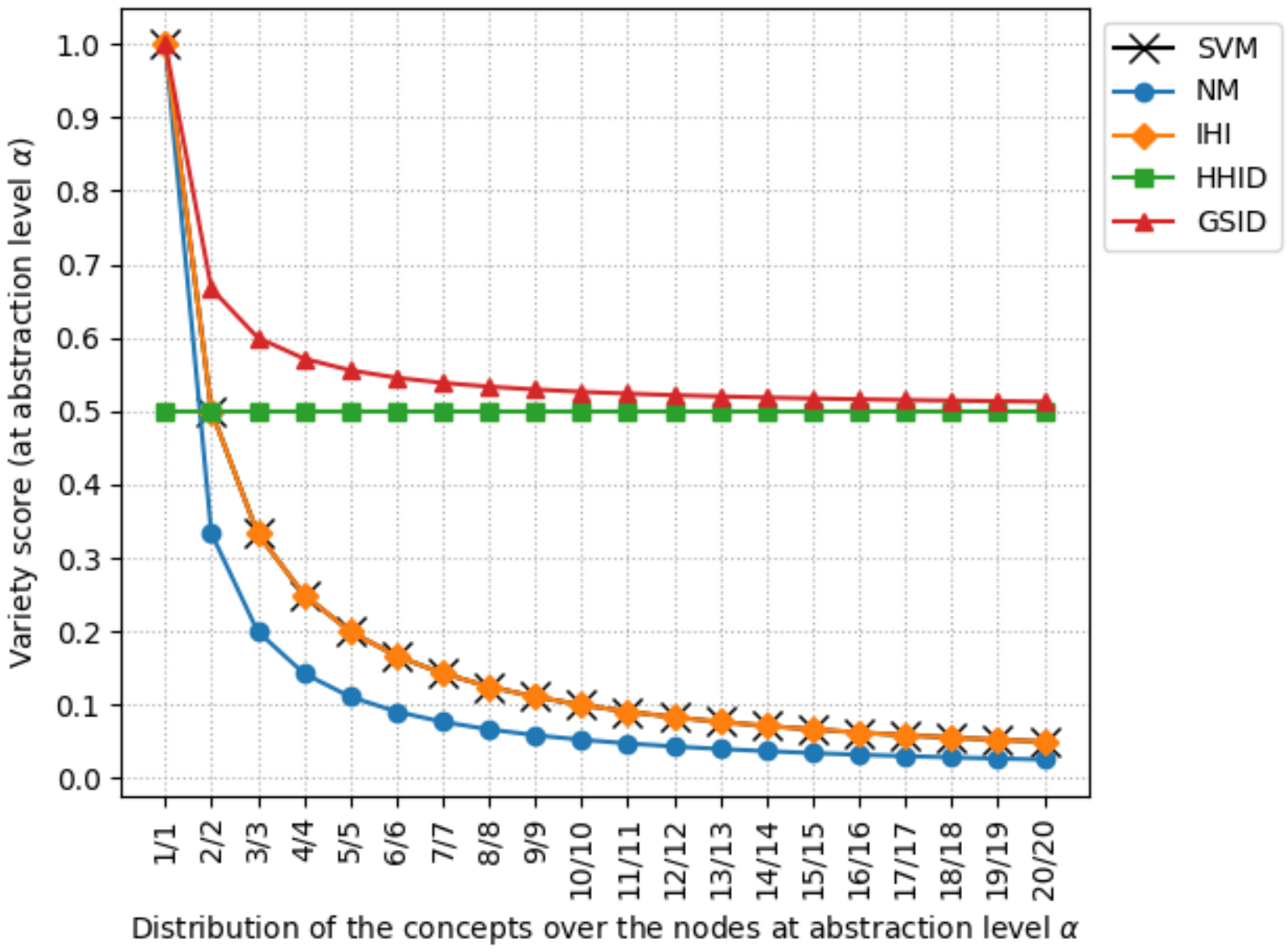

**Figure 8.** Illustration of the Test Case II.

### 3.3 Rationale for Introducing a New Distance-Based Variety Metric

Test Cases I and II suggest that for genealogy tree-based variety assessment, the 'bias-corrected' GSID is a better metric for assessing the variety of a concept space compared to other existing metrics. Additionally, Test Case II implies that in the future, researchers should consider an 'unbiased' estimator of any existing entropy-based measure, such as the Sharma-Mittal entropy (Ahmed et al., 2021), while proposing a new metric.

However, a major assumption for all the existing tree-based variety assessment approaches is that all the ideas in the idea space are considered equally distant. Let's consider an example where three different types of clutches are there in a concept space: plate clutch ($C_1$), centrifugal clutch ($C_2$), and electromagnetic clutch ($C_3$). If we consider the idea space for 'driving input' (representing a design attribute or an abstraction level), then the three different ideas with respect to $C_1$, $C_2$, and $C_3$ are 'spring force,' 'centrifugal force,' and 'electric current,' respectively. In this case, a domain expert may find 'spring force' and 'centrifugal force' to be more similar to each other compared to 'electric current.' Thus, constructing a tree-like structure or grouping ideas in distinct nodes becomes difficult when the distance variable between two ideas is continuous instead of binary. In such cases, none of the existing metrics are applicable as a measure of variety. The same limitation of existing metrics was also pointed by Ahmed et al. (2021).

From Equations 2, 3 and 5, it can be observed that SVS, NM and IHI metrics use the weight of each function (that a design concept needs to satisfy) to assess variety. However, these function weights relate to the value or usefulness of a concept, rather than its variety. Variety should indicate how different one concept is from others, not how good or important the delivered functions are relative to one another. Consequently, using function weights conflates the assessment of concept variety with evaluations of concept quality.

The existing variety metrics discussed in this paper have used four abstraction or hierarchical levels (physical principle, working principle, embodiment and detail) to represent a concept. These four levels were first introduced by Shah et al. (2003). However, the distinction between a physical principle and a working principle remains unclear, and there is a lack of comprehensive guidelines to break a design concept into these four abstraction levels. For instance, Ramachandran et al. (2018) reported their struggle to define the physical and working principles appropriately while creating genealogy trees for the design concepts. Hence, it is up to the designers to carefully describe the principles of the design concepts to ensure repeatability of their variety scores. The problem becomes more evident for complex design concepts which may use multiple physical principles at different operating states to satisfy an intended function (Majumder et al. 2023). For example, a hair dryer can have two operating states: $ON_1$ and $ON_2$. At $ON_1$, it uses two physical principles, 'heating' and 'blowing' of air, whereas, at $ON_2$, it only uses 'blowing' of air to satisfy the intended function. Hence, it is important to choose an appropriate knowledge representation scheme which can describe a complex design concept more comprehensively.

Our work addresses the above issues from three different aspects:

1. *Eliminating the requirement of genealogy tree-based representation:* We eliminate the requirement of genealogy tree-based representation of the concept space and provide a new distance-based variety metric, based on Rao's quadratic diversity index (Rao 1982), that is applicable to measure the variety of a concept space irrespective of which abstraction levels and how many of them are considered while representing the concepts.
2. *Providing a flexible framework for representing and comparing concepts:* In addition to introducing the new metric, we offer a framework for assessing variety that allows design concepts to be described using any preferred representation scheme and level of abstraction. Within this framework, the distance between two concepts at an abstraction level $\alpha$ is measured by analyzing the semantic similarity between their respective descriptions, and all pairwise distances are stored in a distance matrix. This flexibility enables designers to work with representations that best suit their domain or design problem context.
3. *Developing a tool to support the variety assessment process:* A software tool – 'VariAnT' (Variety Assessment Tool) – has also been developed to automate the variety assessment process for a given design concept space.

## 4 A Prescriptive Framework for Assessing Variety

The overall procedural breakdown of the prescriptive variety assessment framework is depicted in **Figure 9**. In the following part of this section, we elaborate on the framework. First, we present a knowledge representation scheme for storing the design concepts into a structured data frame. Next, we demonstrate a method for measuring the distance between two design concepts by comparing the textual descriptions across different abstraction levels, utilising a state-of-the-art Natural Language Processing (NLP) technique. Following this, we propose a new variety metric based on an unbiased estimate of Rao's Quadratic Diversity Index (Rao 1982) and introduce the 'VariAnT' tool, which embodies the proposed framework. Finally, we provide an example to demonstrate the variety calculation using the new framework.

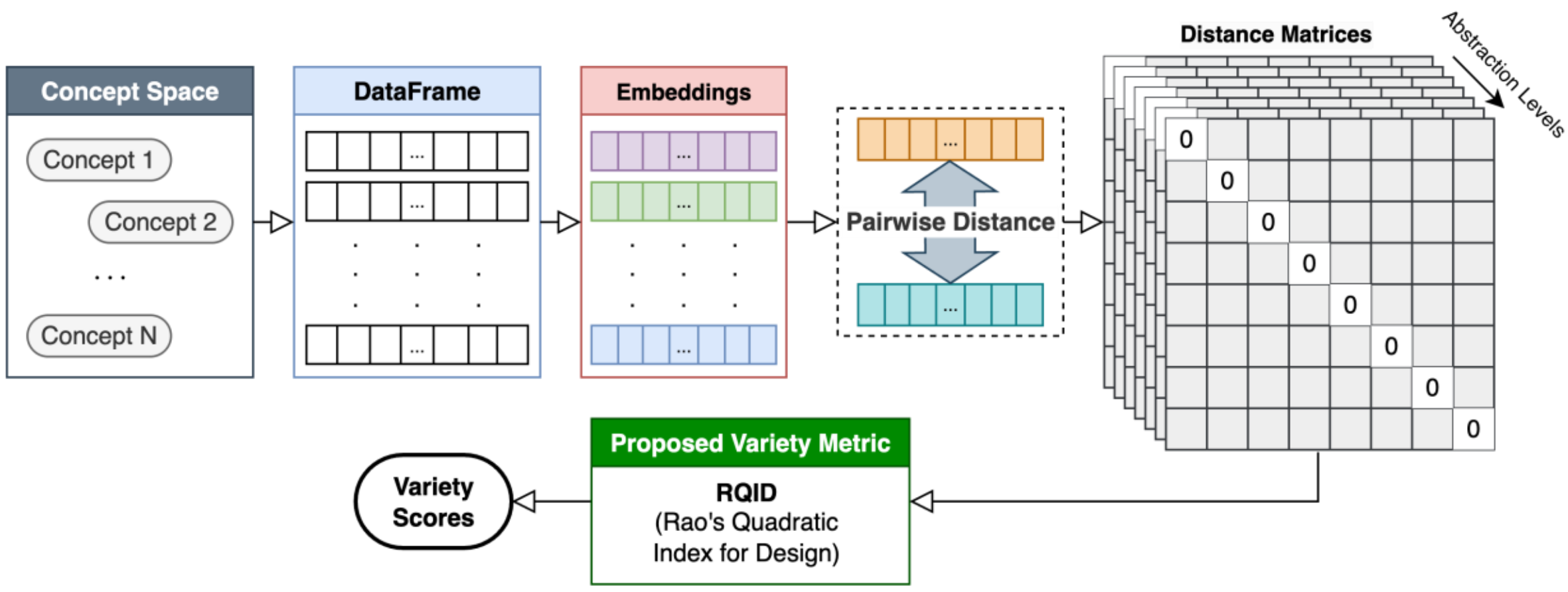

**Figure 9.** The overall procedural breakdown of the proposed variety assessment framework.

### 4.1 Knowledge Representation

#### 4.1.1 Different Representation Schemes and Complexities

Design concepts can be represented at different levels of abstraction, and several modelling schemes in the literature support such multi-level descriptions. For example, the Function-Behaviour-Structure (FBS) model (Gero & Kannengiesser, 2004) represents concepts through the relationships between what a system is intended to do (function), how it does it (behaviour), and what it is composed of (structure). Similarly, Chakrabarti et al. (2005) introduced SAPPhIRE, a causality model, that describes systems across a richer set of abstraction levels, namely, States, Actions, Parts, Phenomena, Inputs, oRgans, and Effects. The interrelationships among these constructs, as depicted in **Figure 10**a, can be summarised as follows: Parts, which encompass physical components and interfaces, play a crucial role in the formation of oRgans, which represent the properties and conditions of a system and its environment necessary for the interaction. oRgans, in conjunction with Input(s) in the form of material, energy, or information, collectively trigger physical Effect(s), which subsequently give rise to physical Phenomena and induce a State change in the system. This State change is further interpreted as Action(s), an abstract description of an interaction that can serve as an input or create/activate (new) Part(s). **Figure 10**b shows an example where the SAPPhIRE model explains how a hot body cools down in the presence of a surrounding fluid medium.

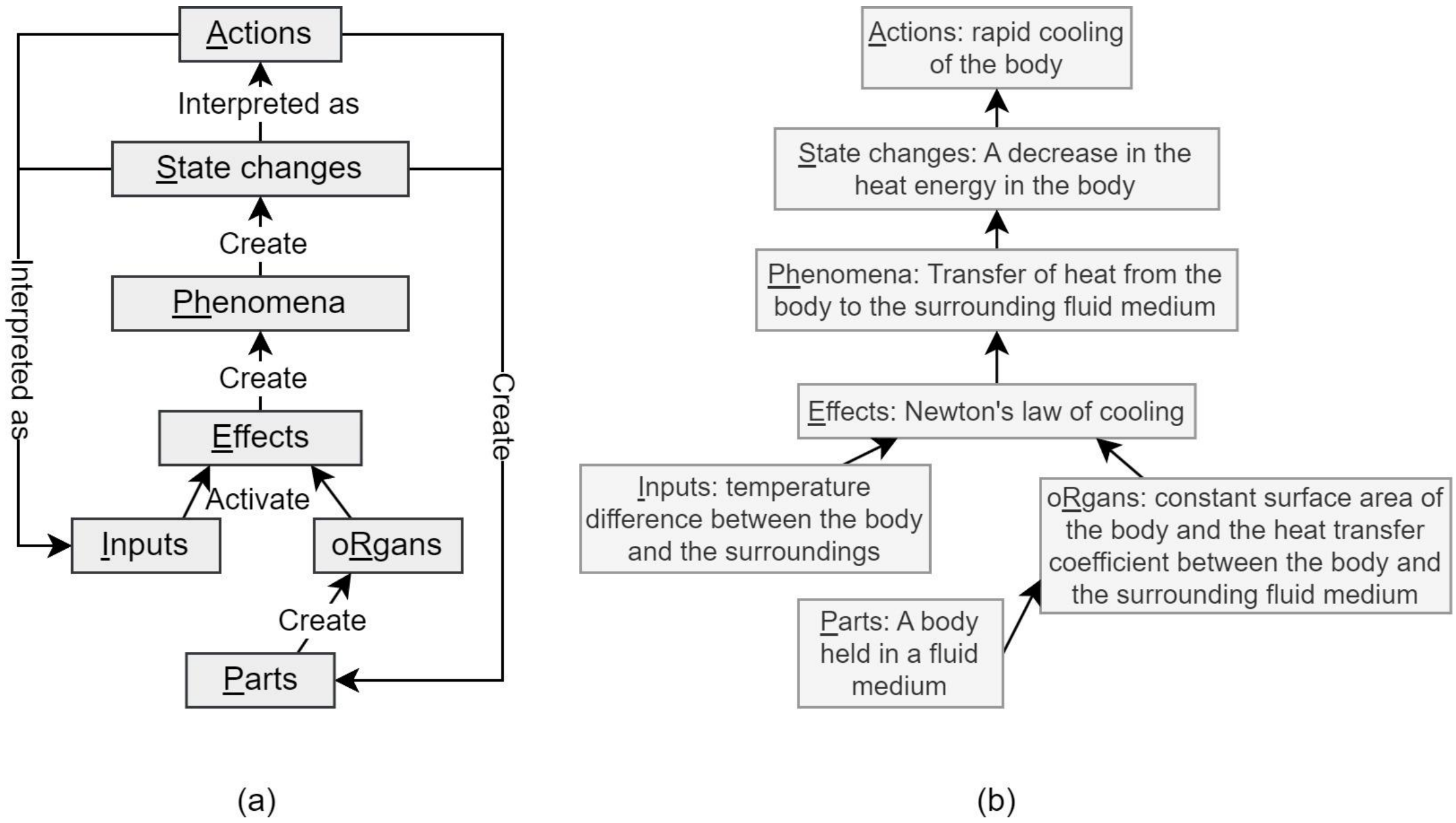

**Figure 10.** (a) The SAPPhIRE model of causality (Chakrabarti et al. 2005), (b) An example SAPPhIRE model explaining how a hot body cools down (Srinivasan & Chakrabarti, 2009).

In comparison to the FBS model, in SAPPhIRE, the 'action' encompasses the notion of 'function,' while 'parts' can be understood as 'structure.' The remaining constructs of SAPPhIRE collectively contribute to a comprehensive depiction of 'behaviour.' Many systems can be represented using a single instance of a SAPPhIRE model. However, when dealing with complex systems that necessitate more detailed descriptions, multiple SAPPhIRE models are required for representation. By using multiple instances of the SAPPhIRE, Siddharth et al. (2018) created causal chains to represent the functioning of a complex system, which provides elements of description that are absent in other existing models. In our past work (Majumder et al., 2023), we found that the majority of existing models lack an explanation of how to model the functionalities of a system with multiple operating states (also referred to as multi-state systems), and the models that do consider multiple operating states do not explicitly explain a method to capture the underlying causal relationships inherent within the system. To address these issues, we proposed an integrated function modelling approach using SAPPhIRE as the basis (Majumder et al., 2023). Broadly, the proposed approach uses the model of a 'transformation system' (Hubka & Eder, 2012) to identify the 'technical process,' the 'operand(s)' (the entity undergoing transformation), and the 'technical system' (or the 'operator') responsible for driving the 'technical process.' Then, the causal processes associated with the 'transformation system' are explained with multiple instances of the SAPPhIRE model. In the past, researchers also reported the use of SAPPhIRE abstraction levels in synthesising design concepts (Srinivasan & Chakrabarti, 2009; Bhatt et al., 2021; Trollman et al., 2023). For example, Trollman et al. (2023) employed the SAPPhIRE model to conceptualise different strategies for countering speculative wheat market fluctuations. Additionally, multiple software tools, such as IDEA-INSPIRE (Chakrabarti et al., 2017) and OPAL (Peters et al., 2021), were developed by researchers that utilise the SAPPhIRE model to create a database of existing design concepts that can be further utilized as design stimuli acting as a design-by-analogy tool. The SAPPhIRE model has also been used to capture the knowledge obtained during the testing phases of product design, such as the potential failure modes (Siddharth et al., 2020).

The above discussion motivates the need for a representation approach that supports multiple descriptive instances, regardless of the specific modelling framework used. Each instance corresponds to a chosen abstraction level or modelling construct (e.g., function, behaviour, structure, input, etc.), and a concept may require one or many such instances depending on its complexity. The framework introduced in this paper operates on these instances without imposing any specific modelling scheme, allowing designers to choose the representation that best suits their domain or design problem.

#### 4.1.2 Concept Space Data Frame

Assuming that we already have explored a concept space $C$ – comprised of $N$ concepts – for a given design problem, each concept in the concept space can be represented with a single instance or a set of instances. For the $i^{th}$ concept ($C_i$), the associated set of instances can be represented as follows:

$$S_i = \{S_i^{\alpha} \mid \alpha \in 1,2, \dots, m\} \tag{10}$$

Where α takes values ranging from 1 to $m$, corresponding to different levels of abstraction. For example, if we consider a SAPPhIRE model, α takes values ranging from 1 to 7, as shown in **Figure 11**. Here, the concept $C_i$ consists of $k_i$ instances of SAPPhIRE, therefore, a list of data or ideas ($S_i^{\alpha}$) at an abstraction level α can be shown as in Equation 11, where $e_i^{\alpha k}$ denotes a construct of the $k^{th}$ instance of the representation model at abstraction level α.

$$S_i^{\alpha} = \{e_i^{\alpha k} \mid k \in 1,2, \dots, k_i\} \tag{11}$$

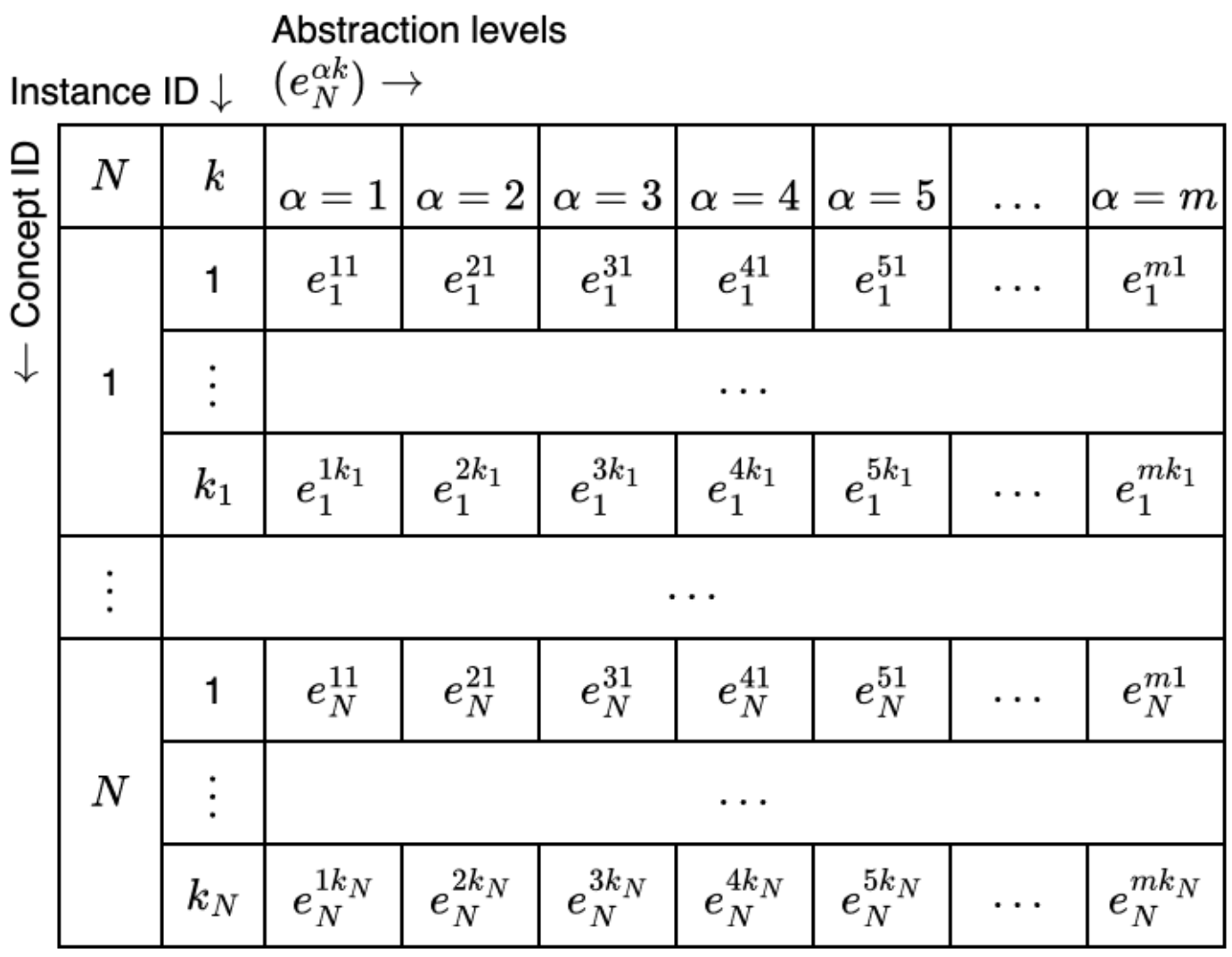

**Figure 11.** Representation of the concept space in a data frame.

### 4.2 Vector Encoding

In this step, the text data stored in the data frame is converted into numerical vectors that can be used to calculate the semantic similarity between the textual data of two concepts at a particular level of abstraction. For each concept, first, we convert the list of strings, $S_i^{\alpha}$, into a single string, $L_i^{\alpha}$ using a concatenation operation, denoted by '$\oplus$' in Equation 12.

$$L_i^{\alpha} = e_i^{\alpha 1} \oplus e_i^{\alpha 2} \oplus \dots \oplus e_i^{\alpha k_i} \tag{12}$$

Next, we consider two such strings: $L_i^{\alpha}$ and $L_j^{\alpha}$ that are generated from the dataset of the $i^{th}$ concept and $j^{th}$ concept at abstraction level α. Then, the strings $L_i^{\alpha}$ and $L_j^{\alpha}$ are encoded as vectors denoted by $\overrightarrow{A_i^{\alpha}}$ and $\overrightarrow{A_j^{\alpha}}$, respectively.

We employ a Sentence-BERT or S-BERT (Reimers & Gurevych, 2019) to generate the text embeddings in our proposed framework. In recent past, S-BERT was widely used across different design applications for embedding natural language texts and evaluating the semantic similarity. For example, Singh and Chakrabarti (2025) used these embeddings to assess the creative potential of design problems, and Wilson and Yang (2025) have utilized similar embeddings to evaluate semantic closeness between human and AI-generated design reasonings. In our current version of the proposed variety assessment tool, we use an open-source, state-of-the-art and lightweight pre-trained sentence transformer model, 'all-MiniLM-L6-v2', provided by HuggingFace[1]. This model maps the text to a 384-dimensional dense vector space and is widely used for NLP tasks like clustering, semantic search, semantic similarity, etc. We may also employ other embedding models, such as OpenAI embeddings[2], universal-sentence-encoder[3], etc., providing access to high-quality, pre-trained language models for various NLP tasks.

### 4.3 Distance Matrix

The distance matrix is a $N \times N$ matrix representing all the pairwise distances (or dissimilarity values) between the design concepts. Once we get the vector embeddings for the $i^{th}$ and $j^{th}$ concept at an abstraction level α, the distance $d_{ij}^{\alpha}$ between $\overrightarrow{A_i^{\alpha}}$ and $\overrightarrow{A_j^{\alpha}}$ is calculated as:

---

[1] https://huggingface.co/sentence-transformers/all-MiniLM-L6-v2
[2] https://platform.openai.com/docs/guides/embeddings/embedding-models
[3] https://pypi.org/project/spacy-universal-sentence-encoder/

$$d_{ij}^{\alpha} = 1 - sim(\overrightarrow{A_i^{\alpha}}, \overrightarrow{A_j^{\alpha}}),\ sim(\overrightarrow{A_i^{\alpha}}, \overrightarrow{A_j^{\alpha}}) = \frac{\overrightarrow{A_i^{\alpha}} \cdot \overrightarrow{A_j^{\alpha}}}{|\overrightarrow{A_i^{\alpha}}||\overrightarrow{A_j^{\alpha}}|} \tag{13}$$

Where, $sim(\overrightarrow{A_i^{\alpha}}, \overrightarrow{A_j^{\alpha}})$ measures the cosine similarity by comparing the orientation of two vectors in a high-dimensional abstract space (Nandy et al., 2022). Further, the distance value $d_{ij}^{\alpha}$ is calculated by subtracting the cosine similarity value from 1. The distance between any two concepts is associative, i.e., $d_{ij}^{\alpha} = d_{ji}^{\alpha}$, and for $i = j$, $d_{ij}^{\alpha} = 0$.

### 4.4 Obtaining Variety Scores

The variety score of the $i^{th}$ concept $(C_i)$ at an abstraction level α becomes the average distance of the $i^{th}$ concept from the other $(N-1)$ concepts in that concept space $(C)$. Therefore,

$$V_i^{\alpha} = \frac{\sum_{j=1}^{N} d_{ij}^{\alpha}}{(N-1)},\ for\ \ N \geq 2 \tag{14}$$

The variety score of a concept space $(C)$ at an abstraction level α becomes the average variety score of all concepts (i.e., $V_i^{\alpha}$) in that concept space, as shown in Equation 15.

$$V^{\alpha}(C) = \frac{\sum_{i,j=1}^{N} d_{ij}^{\alpha}}{N(N-1)},\ for\ \ N \geq 2 \tag{15}$$

The expression of $V^{\alpha}$ in Equation 15, derived using simple heuristics, is equivalent to an unbiased estimator of Rao's Quadratic Diversity Index (Rao 1982) - one of the widely used measures of ecological diversity (Pavoine et al. 2005, Daly et al. 2018). Henceforth, Equation 15 is referred to as Rao's Quadratic Index for Design or RQID.

RQID is valid when the distance values, $d_{ij}^{\alpha}$, satisfy the following properties: symmetry ($d_{ij}^{\alpha} = d_{ji}^{\alpha}$), zero self-distance ($d_{ii}^{\alpha} = 0$), triangle inequality ($d_{ij}^{\alpha} \leq\ d_{ik}^{\alpha} + d_{kj}^{\alpha}$), non-negativity ($d_{ij}^{\alpha} \geq 0$), and standardization of distances to the interval $0 \leq d_{ij}^{\alpha} \leq 1$. (Ricotta and Szeidl, 2006)

Note that RQID reduces to the GSID (Equation 9) in the case where all $d_{ij}^{\alpha}$ takes the value 0 or 1, i.e., the distance variable between two ideas is binary-valued. Detailed mathematical proof regarding the unbiased estimate of Rao's Quadratic Diversity Index and its relationship with the Gini-Simpson Index can be found in Chen et al. (2018).

Now, considering all abstraction levels (i.e., α=1...m), the weighted average variety of an individual concept $V(C_i)$ and the concept space $V(C)$ as a whole can be calculated as follows:

$$V(C_i) = \frac{\sum_{\alpha=1}^{m} w_{\alpha} V_i^{\alpha}}{\sum_{\alpha=1}^{m} w_{\alpha}} \tag{16}$$

$$V(C) = \frac{\sum_{\alpha=1}^{m} w_{\alpha} V^{\alpha}(C)}{\sum_{\alpha=1}^{m} w_{\alpha}} \tag{17}$$

Similarly, we can also get a weighted average distance matrix $D(C) = [D_{ij}]_{N \times N}$ for the concept space $C$ considering all abstraction levels, where an element $D_{ij}$ of the matrix $D(C)$ is calculated as follows:

$$D_{ij} = \frac{\sum_{\alpha=1}^{m} w_{\alpha} d_{ij}^{\alpha}}{\sum_{\alpha=1}^{m} w_{\alpha}} \tag{18}$$

The weights ($w_{\alpha}$) of different abstraction levels can be assigned to reflect the relative importance of each level. Higher weights can be used for abstraction levels where differences between concepts are considered more significant, resulting in a higher variety score for the concept space. In practice, these weights can be defined by the user in the database sheet according to the design problem and the intended emphasis on particular abstraction levels.However, finding an optimised set of weights requires further research, which is outside this article's scope.

### 4.5 Computer Implementation

A software tool – called VariAnT (Variety Assessment Tool) – has been developed to automate the assessment process. The tool's Graphical User Interface (GUI) is created with the PySimpleGUI[4] Python library and is shown in **Figure 12**. The GUI consists of the following six panels:

1. The user can create an Excel file consisting of the concept space information (i.e., $S_i, i\ =\ 1 \ldots N$) of all the concepts (i.e., $C_i, i\ =\ 1 \ldots N$) of a concept space (i.e., $C$ with $N$ concepts). This file can be imported into the tool data frame using the *Import Data* panel. In the current implementation, the weights for each abstraction level must be specified in brackets as numeric values alongside the abstraction level headings (e.g., 'physical principle (10)').
2. Once the data is imported into the tool, the user can use the *Concept space* panel to display the current data frame. This panel displays the data frame where each row corresponds to a particular description (i.e., $e_i^{\alpha k}$) of a concept $C_i$.

[4] https://www.pysimplegui.org/

3. VariAnT allows the user to choose a vector encoding method, e.g., S-BERT, as discussed in Section 4.2. The user can initiate the assessment process by clicking the *Calculate Variety* button. The tool enables users to apply uniform weights or customise the weight values for different abstraction levels through a checkbox.
4. Once the calculations are done, the variety score of the concept space, i.e., $V\ (C)$ (see Equation 17), is displayed in the *Results* panel. In addition to the $V\ (C)$ score, the tool generates four different plots to provide more insights to the user. These plots are discussed in detail in the following section.

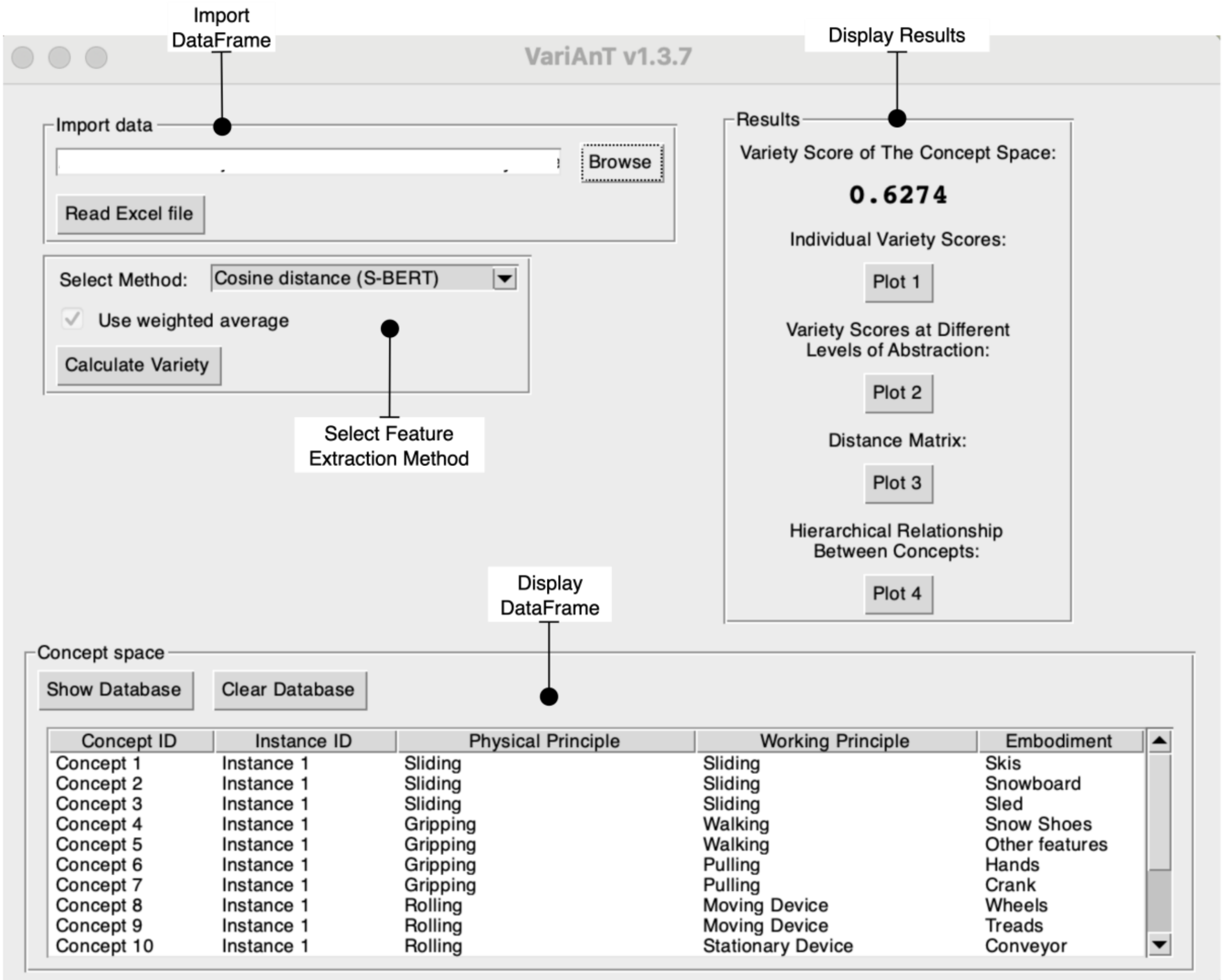

**Figure 12.** VariAnT user interface.

## 5 Demonstration and Evaluation of the Proposed Framework with Example Cases

### 5.1 Variety Assessment of Design Concepts Represented using SAPPhIRE Abstraction Levels

To demonstrate the proposed variety assessment framework, a synthetic concept space ($C^W$ , $N\ =\ 4$) is considered, which consists of 4 concepts generated using the web interface of ChatGPT[5]. We have used the GPT-4o model with default settings. The concepts were generated in response to the following design problem: 'How to boil water?' The model was asked to provide four concepts and explicitly describe them using seven abstraction levels of SAPPhIRE, adhering to the definitions of SAPPhIRE constructs. The prompting strategy used for this purpose is outside this paper's scope and is discussed elsewhere (Majumder et al., 2025). The concepts generated and corresponding SAPPhIRE constructs are given in Table 1. Now, given the SAPPhIRE descriptions of the four concepts of the concept space $C^W$, the variety assessment is carried out using the software tool. The following text embedding model is used: 'all-MiniLM-L6-v2'; and the weights ($w_\alpha$) of different abstraction levels are arbitrarily assigned as: $w_1 = 7$ (Actions), $w_2 = 6$ (States), $w_3 = 5$ (Inputs), $w_4 = 4$ (Phenomena), $w_5 = 3$ (Effects), $w_6 = 2$ (oRgans), $w_7 = 1$ (Parts). The weightage values are adapted from Srinivasan & Chakrabarti (2010a). The rationale is to obtain a higher variety score for a given concept space where the concepts differ at a higher abstraction level. The results obtained from the tool are discussed as follows.

[5] https://chatgpt.com/

**Table 1.** Design concepts and the corresponding SAPPhIRE models generated with the help of ChatGPT (GPT-4o).

| Abstraction Levels | Concept 1: Electric Kettle | Concept 2: Gas Stove with Kettle | Concept 3: Solar Water Heater | Concept 4: Friction Heater |
|---|---|---|---|---|
| Actions | Boiling of Water | Boiling of Water | Boiling of Water | Water becomes warm |
| (change of) States | Initial State: Cold Water (low temperature), Final State: Hot Water (high temperature) | Initial State: Cold Water (low temperature), Final State: Hot Water (high temperature) | Initial State: Cold Water (low temperature), Final State: Hot Water (high temperature) | Initial State: Cool Water (moderate temperature), Final State: Warm Water (higher temperature) |
| Phenomena | Conversion of electrical energy to heat energy, Transfer of heat energy to water | Combustion of Gas (conversion of chemical energy to heat energy), Transfer of heat from flame to kettle, Transfer of heat from kettle to water | Conversion of solar energy to heat energy, Transfer of heat to water in tank | Conversion of mechanical energy to heat energy through friction, Transfer of heat from friction rods to water |
| Effects | Joule Heating, Heat Conduction | Chemical reaction between gas and oxygen producing heat, Heat Conduction | Photovoltaic Effect, Heat Transfer | Frictional Heating |
| Inputs | Electrical energy (from power supply) | Natural Gas (fuel) | Solar energy (sunlight) | Mechanical energy (from manual rotation of the crank handle) |
| oRgans | Conductive material for the heating element, Contact between the heating element and water container, Insulated environment to prevent heat loss, Closed system (lid closed) | Combustible Gas, Oxygen (from air), Metallic Kettle (good conductor of heat), Stable Pot Stand, Controlled Environment (regulated flame) | Presence of Photovoltaic Cells, Conductive Pipes, Insulated Tank, Availability of Sunlight | High-friction Material for Rods, Smooth Bearings to ensure easy rotation, Water-tight Container, Insulated Environment to prevent heat loss, Constant Manual Input (consistent rotation) |
| Parts | Heating Element, Water Container, Power Cord, Lid, Base Plate, Switch, Insulation | Gas Burner, Kettle, Gas Supply Pipe, Knob (for control), Flame Ignition Mechanism, Pot Stand | Solar Panels, Water Tank, Pipes, Insulation, Heat Exchanger, Temperature Sensor | Friction Rods, Water Container, Rotating Shaft, Bearings, Crank Handle |

### 5.1.1 Variety Assessment Results

The variety score of the concept space as a whole is calculated as $V\,(C^W) = 0.378$. The individual variety scores ($V\,(C_i), i = 1 \ldots 4$) are exported as a bar chart from the tool, as shown in **Figure 13**a. Here, the Friction Heater's individual variety score, $V\,(C_4^W) = 0.46$, is the highest compared to other concepts. This interpretation aligns with the notion of 'a-posteriori' where a concept's distinctiveness is defined relative to the set it belongs to, as discussed in Fiorineschi et al. (2022).

**Figure 13**b shows the variety score of the concept space ($C^W$) at an abstraction level $\alpha$, i.e., $V^\alpha(C^W)$ (see Equation 15). In this figure, $V^\alpha(C^W)$ $(= \overline{V_i^\alpha})$ score, depicted in the box plot, is represented by a horizontal orange line inside the box, providing a visual indication of the central tendency of the $V_i^\alpha$ scores (see Equation 14). Here, outliers are shown in red circles. At a particular abstraction level, these outlier values indicate which concept(s) significantly varied from the other concepts. For example, **Figure 13**b shows that at the State change level, $V_i^{\alpha=states}$ score of the Friction Heater is higher compared to the other 3 concepts. Srinivasan & Chakrabarti (2010a) conducted observational studies involving an individual designer in each design session to solve a conceptual design problem. The objective was to test whether "an increase in the size and variety of ideas used while designing should enhance the variety of concepts produced, leading to an increase in the novelty of the concept space." They concluded that a more diverse set of ideas generated at a higher abstraction level leads to a greater chance of producing a newer concept. Thus, examining the variety score across different levels of abstraction, as shown in **Figure 13**b, may provide valuable insights into the idea-space exploration. For example, while synthesising concepts, a designer or design team can focus on the abstraction level with a low variety score and try to diversify the ideas in that particular idea space.

**Figure 13**c shows a heatmap of the weighted average distances ($D_{ij}$) between each pair of concepts ($C_i^W, C_j^W$) where, $i, j = 1 \ldots 4$. Using these pairwise distance values, the tool can perform clustering where the user defines the number of clusters. Once the number of clusters is defined, the tool implements the K-means clustering algorithm and prints the resulting cluster labels. Additionally, it generates a dendrogram from the clustering results, as shown in **Figure 13**d. This plot helps to visualise sets of concepts that are similar to each other. Researchers argued that it is easier to explore a large concept space meaningfully when designers browse a clustered concept space because they only need to go through a few concepts from each cluster to get a fair overview, instead of going through every concept (Langdon & Chakrabarti, 1999).

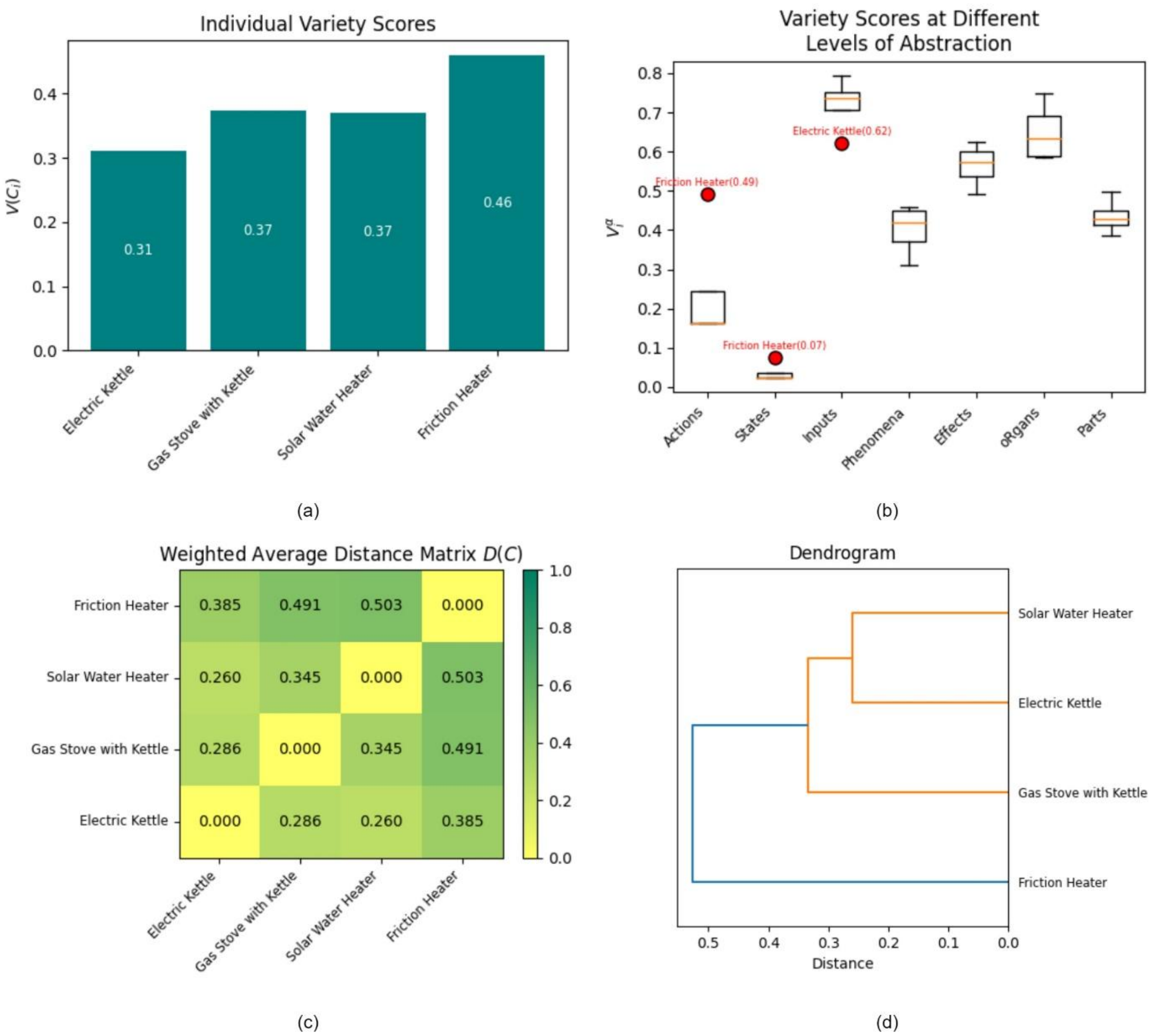

**Figure 13.** Results obtained for the concept space $C^W$: (a) Individual variety scores; (b) Variety scores of the concept space at different levels of abstraction; (c) The weighted average distances between each pair of concepts; (d) Dendrogram generated from the clustering results.

## 5.2 Comparative Evaluation with Existing Methods

Here, we evaluate the proposed framework against the existing methods discussed in Section 2.2. To demonstrate practical applicability - unlike the synthetic concept space used previously - we employ ideas generated from a real ideation exercise reported by Henderson et al. (2017). The problem brief, discussed in detail in their study, concerns generating concepts for transportation in snowy environments. The concepts used in this evaluation are listed in Table 2.

**Table 2.** Design concepts partially taken from Henderson et al. (2017).

| Concepts | Abstraction Levels | | |
|---|---|---|---|
| | Physical principle | Working principle | Embodiment |
| $C_1^T$ | Sliding | Sliding | Skis |
| $C_2^T$ | Sliding | Sliding | Snowboard |
| $C_3^T$ | Sliding | Sliding | Sled |
| $C_4^T$ | Gripping | Walking | Snowshoes |
| $C_5^T$ | Gripping | Walking | Other features |
| $C_6^T$ | Gripping | Pulling | Hands |
| $C_7^T$ | Gripping | Pulling | Crank |
| $C_8^T$ | Rolling | Moving Device | Wheels |
| $C_9^T$ | Rolling | Moving Device | Treads |
| $C_{10}^T$ | Rolling | Stationary Device | Conveyor |
| $C_{11}^T$ | Air Pressure | Wind | Sail |
| $C_{12}^T$ | Air Pressure | Gliding | Airplane |
| $C_{13}^T$ | Air Pressure | Fan | Hovercraft |
| $C_{14}^T$ | Air Pressure | Fan | Helicopter |
| $C_{15}^T$ | Air Pressure | Fan | Fan Boat |
| $C_{16}^T$ | Air Pressure | Propulsion | Jet |
| $C_{17}^T$ | Hopping | Spring lift | Bouncy shoes |
| $C_{18}^T$ | Hopping | Spring lift | Jump Platform |
| $C_{19}^T$ | Hopping | Spring lift | Elastic Exoskeleton |
| $C_{20}^T$ | Magnetic Repulsion | Cushion | Repulsive Shoes |
| $C_{21}^T$ | Magnetic Repulsion | Cushion | Other features |
| $C_{22}^T$ | Magnetic Repulsion | Guide Rail | Maglev Train |
| $C_{23}^T$ | Magnetic Repulsion | Guide Rail | Hover sled |

The first sixteen concepts in Table 2, denoted as $C_i^T, i = 1...16$, are taken directly from Henerson et al. (2017). To facilitate comparative evaluation, seven additional concepts, i.e., $C_i^T, i = 17...23$, were created and appended to the original set.

Importantly, if we divide the concept space $C^T$ into two sets: Set P ($C_i^T, i = 1...16$) and Set Q ($C_i^T, i = 8...23$), both the sets result in identical genealogy trees (refer to section 2.1). Consequently, all the existing variety assessment methods, discussed in Section 2.2, would assign identical variety scores to both sets. However, it is evident from the underlying data that, at the level of physical principles, the ideas in Set Q are conceptually more heterogeneous than those in Set P. A reliable variety assessment method should be capable of capturing this distinction. Without this sensitivity, variety would collapse into a mere count of ideas, capturing only fluency, and the distinction between fluency and variety as two separate creativity measures would effectively disappear. Therefore, detecting conceptual distance at different abstraction levels is fundamental to preserving the theoretical integrity and practical usefulness of variety as a metric.

When the data for Sets P and Q are processed using the proposed tool, the resulting variety scores are 0.627 and 0.692, respectively. These results were obtained using the 'all-MiniLM-L6-v2' text embedding model and the following user-defined abstraction-level weights ($w_\alpha$): $w_1 = 10$ (physical principle), $w_2 = 6$ (working principle), $w_3 = 3$ (embodiment). The increase in the score for Set Q demonstrates the sensitivity of the proposed framework to conceptual differences at higher abstraction levels and also illustrates its adaptability to different knowledge representation schemes.

### 5.3 Limitations of the Proposed Framework and Tool

Following are some of the limitations of the proposed framework and the current implementation of the 'VariAnT' tool:

1. The current algorithm has a computational complexity of $O(N^2)$, where $N$ is the number of concepts. As the number of concepts or abstraction levels increases, the computation time may grow significantly. However, this problem can be mitigated by utilizing parallel computing could mitigate, which is currently not implemented in the tool.
2. Generating text embedding can be time-consuming or costly, depending on the embedding model used and the computational resources available. Larger models or embeddings with higher dimensionality may require longer processing times.
3. The current tool does not include methods for optimizing the selection of concepts to maximize variety. For example, in a design-by-analogy tool, e.g., IDEA-INSPIRE (Chakrabarti et al., 2017), one could identify $n$ stimuli from a total of $N$ concepts such that the variety of the selected subset is maximized. Similar functionality was addressed by Ahmed et al., (2021) where greedy algorithm was implemented, and the same can be implemented in 'VariAnT' as well in future.
4. The tool currently operates only on concept spaces represented in textual form. Extending the framework to handle sketches, images, or CAD models would require the use of image embedding, or 3D feature embedding methods, but the underlying framework would remain the same. Implementation for non-textual inputs is planned for future work.

5. The accuracy of the variety assessment depends on the quality and consistency of the textual descriptions of concepts. In the demonstrations presented, concept descriptions were assumed to be accurate and consistent. Issues such as spelling errors, factual inaccuracies, or inconsistent descriptions are not automatically corrected by the tool and rely on careful user input.
6. The usability of VariAnT has not been evaluated in real-world design scenarios with designers. This aspect is beyond the scope of the current study but represents an important direction for future research.

## 6 Conclusions and Future Work

This paper reviewed several existing metrics from the engineering design literature that were proposed to quantify the variety of design concept spaces. It was found that most of the existing variety metrics employ a genealogy tree-based approach. Two test cases were conducted to highlight the limitations of these existing variety metrics. The metrics were evaluated in terms of accuracy and sensitivity. Here, the term 'accuracy' denotes the validity of a metric; 'sensitivity' indicates the ability of a metric to reflect a change in measurement with varying numbers of concepts in a concept space, as well as the distribution of concepts over the nodes of their genealogy tree. Results from two test cases suggested that the 'bias-corrected' GSID (Gini-Simpson Index for Design) was better for assessing the variety of a concept space than the other existing metrics. It was found that a major assumption underlying all these metrics is that the ideas in the idea space are considered equally distant. However, in practice, a real-valued distance can be obtained between two different concepts at a particular abstraction level.

To address this research gap, a new prescriptive framework has been proposed for assessing the variety score of a concept space. One of the distinct features of the proposed framework is the use of the SAPPhIRE model of causality as a knowledge representation scheme, which resolves the problem of representing complex systems. The other significant advantage of the framework is using the RQID (Rao's Quadratic Index for Design) as a variety metric, which enables us to use real-valued pairwise distances instead of creating a genealogy tree. It was found that RQID reduces to the GSID if the distance variable between two ideas is binary-valued, i.e., 0 or 1. Hence, the proposed metric, by default, qualifies for a genealogy tree-based variety assessment. We have also discussed a step-by-step approach to measure the distance between two design concepts by comparing their respective textual descriptions. The proposed variety assessment framework was embodied into a new software tool called 'VariAnT.' The tool provides a GUI and automates the proposed variety assessment process. Finally, the tool was tested with an example concept space, and the results obtained from the tool were discussed.

There are three potential research directions in the future: (a) Besides S-BERT, it is also possible to incorporate several alternative vector encoding methods into the framework, such as OpenAI embeddings, universal-sentence-encoder, etc. Thus, finding the most appropriate method or combination of methods that correlate better with expert variety score ratings requires an empirical investigation; (b) In this work, the weights of different abstraction levels have been assigned arbitrarily, where the highest importance is given to the function followed by behaviour and structure. Therefore, additional optimisation algorithms can be incorporated to optimise the set of weights that can maximise the sensitivity or discriminating power of the proposed framework; (c) Lastly, apart from assessing the variety, the proposed framework could also be integrated with existing 'design-by-analogy' or 'analogy retrieval' tools, such as IDEA-INSPIRE (Chakrabarti et al., 2017), DANE (Vattam et al., 2011), etc., so as to provide insights into the variety of the analogy space retrieved by the tool. The proposed method could also be utilised to control analogical distance while performing a search.

**Competing interests.** The authors declare that they have no known competing financial interests or personal relationships that could have appeared to influence the work reported in this paper.

**Data availability statement.** The authors confirm that all data generated or analysed during this study are included in the paper.